\pdfoutput=1
\documentclass[11pt]{article}

\usepackage[final]{naacl2021}

\usepackage[T1]{fontenc}
\usepackage[utf8]{inputenc}

\usepackage{tikz-qtree}
\usepackage[inline]{enumitem}
\usepackage[pdf]{graphviz}

\usepackage{microtype}
\usepackage[ruled,vlined,linesnumbered]{algorithm2e}
\usepackage{times}
\usepackage{latexsym}
\usepackage{amsmath}
\usepackage{url}
\usepackage{mathrsfs}
\usepackage{amsmath}
\usepackage{tikz}

\usetikzlibrary{shapes}

\usepackage{graphicx}
\graphicspath{ {./figs/} }

\DeclareMathOperator*{\argmax}{arg\,max}

\newcommand\jb[1]{\textcolor{blue}{[JB: #1]}}
\newcommand\oh[1]{\textcolor{red}{[OR: #1]}}

\newcommand\commentout[1]{}
\newcommand\sS{\mathcal{S}}

\newcommand\sE{\mathcal{E}}
\newcommand\sU{\mathcal{U}}
\newcommand\sB{\mathcal{B}}
\newcommand\sT{\mathcal{T}}
\newcommand\sZ{\mathcal{Z}}
\newcommand\bfx{\textbf{x}}
\newcommand\bfs{\textbf{s}}
\newcommand\bfb{\textbf{b}}
\newcommand\bfw{\textbf{w}}
\newcommand\bfe{\textbf{e}}
\newcommand\bfu{\textbf{u}}
\newcommand\bfr{\textbf{r}}
\newcommand\bfz{\textbf{z}}
\newcommand\hash{\text{hash}}

\newcommand\ourparser{\textsc{SmBoP}}
\newcommand\mycomment[1]{}

\title{\ourparser{}: Semi-autoregressive Bottom-up Semantic Parsing}

\author{Ohad Rubin \\
  Tel Aviv University \\
  \texttt{ohadr@mail.tau.ac.il} \\\And
  Jonathan Berant \\
  Tel Aviv University\\
  Allen Institute for AI \\
  \texttt{joberant@cs.tau.ac.il} \\}

\date{}

\begin{document}
\maketitle
\begin{abstract}
The de-facto standard decoding method for semantic parsing in recent years has been to autoregressively decode the abstract syntax tree of the target program using a top-down depth-first traversal. In this work, we propose an alternative approach: a Semi-autoregressive Bottom-up Parser (\ourparser{}) that constructs at decoding step $t$ the top-$K$ sub-trees of height $\leq t$. Our parser enjoys several benefits compared to top-down autoregressive parsing.
From an efficiency perspective, bottom-up parsing allows to decode all sub-trees of a certain height in parallel, leading to logarithmic runtime complexity rather than linear. From a modeling perspective, a bottom-up parser learns representations for meaningful semantic sub-programs at each step, rather than for semantically-vacuous partial trees.
We apply \ourparser{} on \textsc{Spider}, a challenging zero-shot semantic parsing benchmark, and show that \ourparser{}
leads to a 2.2x speed-up in decoding time and
a $\sim$5x speed-up in training time, compared to a semantic parser that uses autoregressive decoding. \ourparser{} obtains 71.1 denotation accuracy on \textsc{Spider}, establishing a new state-of-the-art, and 69.5 exact match, comparable to the 69.6 exact match of the autoregressive \textsc{RAT-SQL+Grappa}.

%obtaining 66.7\% exact match -- the second-best reported result on \textsc{Spider}, only 2.9\% behind \textsc{RAT-SQL+Grappa}.

%Last, \ourparser{} includes Transformer-based layers that contextualize sub-trees with one another, allowing us, unlike traditional beam-search, to score trees conditioned on other trees that have been previously explored.
%we propose \emph{contextualized search}, where at each step all sub-trees are contextualized by one another. This lets the parser score trees based on a global view, where each sub-tree is conditioned on all other sub-trees. 
\end{abstract}

\section{Introduction}

% The goal of zero-shot semantic parsing  is to map language utterances into executable programs in a new environment, or database.
% The standard for this task is encoder decoder models employing  seq2seq to generate the  top-down  derivation of the logical form.
% This work is about generating logical forms in a bottom-up manner using a novel bottom-up parsing algorithm.
% The proposed algorithm decodes in logarithmic time compared to current algorithms which are linear.
% Following previous work
% % \cite{???} 
% we use a intermediate form (RA) and infer an equivalent SQL query in a deterministic fashion.

Semantic parsing, the task of mapping natural language utterances into programs \citep{zelle1996,zettlemoyer2005,clarke2010, liang2011}, has converged in recent years on a standard encoder-decoder architecture. 
Recently, meaningful advances emerged on the encoder side, including developments in Transformer-based architectures \cite{rat-sql} and new pretraining techniques \cite{yin-etal-2020-tabert,herzig-etal-2020-tapas,yu2020grappa,deng2020structure,shi2021learning}.
Conversely, the decoder has remained roughly constant for years, where the abstract syntax tree of the target program is autoregressively decoded in a top-down manner \citep{yin2017syntactic,krishnamurthy2017neural,rabinovich2017abstract}.

Bottom-up decoding in semantic parsing has received little attention \citep{Jianpeng2019,odena2020bustle}. In this work, we propose a bottom-up semantic parser, 
and demonstrate that equipped with recent developments in Transformer-based \citep{vaswani2017} architectures, it offers several advantages. From an efficiency perspective, bottom-up parsing can naturally be done \emph{semi-autoregressively}: at each decoding step $t$, the parser generates \emph{in parallel} the top-$K$ program sub-trees of depth $\leq t$ (akin to beam search).
This leads to runtime complexity that is logarithmic in the tree size, rather than linear, contributing to the rocketing interest in efficient and greener artificial intelligence technologies \cite{Schwartz2020green}.
From a modeling perspective, neural bottom-up parsing provides learned representations for meaningful (and executable) sub-programs, which are sub-trees computed during the search procedure, in contrast to top-down parsing, where hidden states represent partial trees without clear semantics.
%This opens the door to new modeling opportunities. For example, before scoring trees on the beam, we contextualize them with a Transformer, allowing the score of each sub-tree to depend on all other sub-trees it may combine with in future steps.

%Last, working with sub-trees allows us to naturally \emph{contextualize} and \emph{re-rank} trees as part of the model. Specifically, before scoring trees on the beam, they are contextualized with a Transformer, allowing us to globally score each tree conditioned on other trees it may combine with in future steps. Similarly, a Transformer-based re-ranking layer takes all constructed tree representations and re-ranks them. This is in contrast to prior work \citep{goldman2017,bogin2019global,yin2019}, where re-ranking programs was done using a completely separate model. 

%our parser naturally lends itself to the concept of \emph{contextualized beam search}, where before scoring the trees on the beam, they are first contextualized with a Transformer. This allows us to score each sub-tree conditioned on a global view of the entire beam, improving the search quality.

\begin{figure*}
    \centering
    \includegraphics[width=0.75\textwidth]{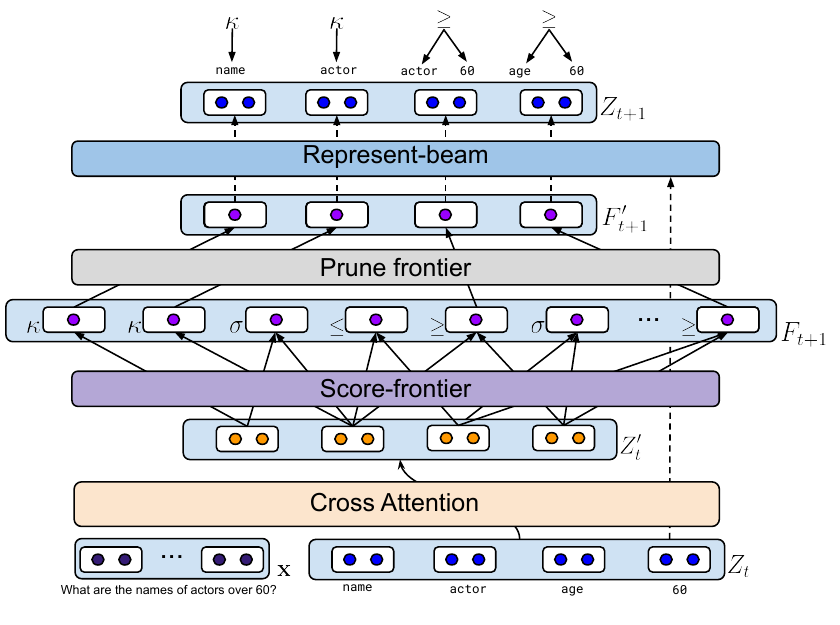}
    \caption{An overview of the decoding procedure of \ourparser{}. $Z_t$ is is the beam at step $t$, $Z'_t$ is the contextualized beam after cross-attention, $F_{t+1}$ is the frontier ($\kappa, \sigma, \geq$ are logical operations applied on trees, as explained below), $F'_{t+1}$ is the pruned frontier, and $Z_{t+1}$ is the new beam. At the top we see the new trees created in this step. For $t=0$ (depicted here), the beam contains the predicted schema constants and DB values. }
    \label{fig:overview}
\end{figure*}

Figure~\ref{fig:overview}  illustrates a single decoding step of our parser. Given a beam $Z_t$ with $K=4$ trees of height $t$ (blue vectors), we use \emph{cross-attention} to contextualize the trees with information from the input question (orange).
%a \emph{beam Transformer} contextualizes the trees on the beam together with the input utterance (orange). 
Then, we score the \emph{frontier}, that is, the set of all trees of height $t+1$ that can be constructed using a grammar from the current beam, and the top-$K$ trees are kept (purple). Last, a representation for each of the new $K$ trees is generated and placed in the new beam $Z_{t+1}$. After $T$ decoding steps,  the parser returns the highest-scoring tree in $Z_T$ that corresponds to a full program.
Because we have gold trees at training time, the entire model is trained jointly using maximum likelihood.

We evaluate our model, \ourparser{}\footnote{Rhymes with `MMMBop'.} (SeMi-autoregressive Bottom-up semantic Parser), on \textsc{Spider} \citep{yu2019}, a challenging zero-shot text-to-SQL dataset. 
We implement the \textsc{RAT-SQL+Grappa} encoder \cite{yu2020grappa}, currently the best model on \textsc{Spider}, and replace the autoregressive decoder with the semi-autoregressive \ourparser{}. \ourparser{} obtains an exact match accuracy of 69.5, comparable to the autoregressive \textsc{RAT-SQL+Grappa} at 69.6 exact match, and to current state-of-the-art at 69.8 exact match \cite{zhao2021gp}, which applies additional pretraining. Moreover, \ourparser{} substantially improves state-of-the-art in denotation accuracy, improving performance from $\text{68.3}\rightarrow\text{71.1}$. Importantly, compared to autoregressive semantic parsing , we observe an average speed-up of 2.2x in decoding time, where for long SQL queries, speed-up is between 5x-6x, and a training speed-up of $\sim$5x.\footnote{ Our code is available at \url{https://github.com/OhadRubin/SmBop}}

\section{Background}
\label{sec:background}
\paragraph{Problem definition}
We focus in this work on text-to-SQL semantic parsing.
% We are given a training set $\{ (x^{(k)},y^{(k)},S^{(k)})  \}$
% \begin{itemize}
%     \item $x^{(k)}$ is a natural language question.
%     \item  $y^{(k)}$ is its translation into Relational Algebra/SQL query.
%     \item $S^{(k)}$  is the schema of the DB where      $y^{(k)}$      is executed.
% \end{itemize}
Given a training set $\{ (x^{(i)},y^{(i)},S^{(i)})\}_{i=1}^N$, where $x^{(i)}$ is an utterance, $y^{(i)}$ is its translation to a SQL query, and $S^{(i)}$ is the schema of the target database (DB), our goal is to learn a model that maps new question-schema pairs $(x, S)$ to the correct SQL query $y$. A DB schema $S$ includes : (a) a set of tables, (b) a set of columns for each table, and (c) a set of foreign key-primary key column pairs describing relations between table columns. Schema tables and columns are termed schema constants, and denoted by $\sS$.
% A Table (or Relation) is  a list of tuples.
% A database schema is a collection of tables that may be connected to each other in some way.
% We call database constants the attributes of each relation/table.
% We can execute SQL query (which is a type of a logical form) on these schema. 
% The theoretical model for SQL is Relational Algebra, which we can transpile to and from.
% A balanced tree is a tree where the depth of every leaf is the same.
\paragraph{RAT-SQL encoder}
This work is focused on decoding, and thus we implement the state-of-the-art RAT-SQL encoder \cite{wang2020}, on top of \textsc{Grappa} \cite{yu2020grappa}, a pre-trained encoder for semantic parsing. 
We now briefly review this encoder for completeness.

The RAT-SQL encoder is based on two main ideas. First, it provides a joint contextualized representation of the utterance and schema. Specifically, the utterance $x$ is concatenated to a linearized form of the schema $S$, and they are passed through a stack of Transformer \citep{vaswani2017} layers. Then, tokens that correspond to a single schema constant are aggregated, which results in a final contextualized representation $(\bfx, \bfs) = (\bfx_1,\dots,\bfx_{|x|},\bfs_1,\dots,\bfs_{|\bfs|})$, where $\bfs_i$ is a vector representing a single schema constant.
This contextualization of $x$ and $S$ leads to better representation and alignment between the utterance and schema.

Second, RAT-SQL uses \emph{relational-aware self-attention} \citep{shaw2018} to encode the structure of the schema and other prior knowledge on relations between encoded tokens. Specifically, given a sequence of token representations $(\bfu_1, \dots, \bfu_{|\bfu|})$, relational-aware self-attention computes a scalar similarity score between pairs of token representations $e_{ij} \propto \bfu_i W_Q (\bfu_j W_K +\bfr_{ij}^K)$. This is identical to standard self-attention ($W_Q$ and $W_K$ are the query and key parameter matrices), except for the term $\bfr_{ij}^K$, which is an embedding that represents a relation between $\bfu_i$ and $\bfu_j$ from a closed set of possible relations. For example, if both tokens correspond to schema tables, an embedding will represent whether there is a primary-foreign key relation between the tables. If one of the tokens is an utterance word and another is a table column, a relation will denote if there is a string match between them. The same principle is also applied for representing the self-attention \emph{values}, where another relation embedding matrix is used. We refer the reader to the RAT-SQL paper for details.
\commentout{
\begin{align*},
    \alpha_{ij} &= \text{softmax}_j\{e_{ij}\} \\
    z_i &= \sum_j \alpha_{ij}(\bfu_j W_V+\bfr_{ij}^V),
\end{align*}
which again is identical to standard self-attention except for the relation embedding $\bfr_{ij}^V$, and $W_V$ is the standard value parameter matrix.
}

Overall, RAT-SQL jointly encodes the utterance, schema, the structure of the schema and alignments between the utterance and schema, and leads to state-of-the-art results in text-to-SQL parsing.

RAT-SQL layers are typically stacked on top of a pre-trained language model, such as BERT \cite{devlin-etal-2019-bert}. In this work, we use \textsc{Grappa} \cite{yu2020grappa}, a recent pre-trained model that has obtained state-of-the-art results in text-to-SQL parsing. \textsc{Grappa} is based on \textsc{RoBERTa} \cite{liu2019roberta}, but is further fine-tuned on synthetically generated utterance-query pairs using an objective for aligning the utterance and query.

%We use a Relation Aware Transformer (RAT) \cite{shaw}, which is an augmented transformer to allow modeling arbitrary relations. Specifically, we use the the question-schema encoding framework of RATSQL\cite{rat-sql} as an encoder and describe it here for defining notation and completeness. A RAT layer consists of the following:
\commentout{
\begin{gather}
    \small
    \begin{aligned}
        e_{ij}^{(h)} &= \frac{\vect{x_i} W_Q^{(h)} (\vect{x_j} W_K^{(h)} +\vect{r_{ij}^K} )^\top}{\sqrt{d_z / H}};
        \alpha_{ij}^{(h)} = softmax_{j} \bigl\{ e_{ij}^{(h)} \bigr\}
        \\
        \vect{z}_i^{(h)} &= \sum_{j=1}^n \alpha_{ij}^{(h)} (\vect{x_j} W_V^{(h)}+\vect{r_{ij}^V}) ;
        \vect{z}_i = [\bigl \vect{z}_i^{(1)}, \cdots, \vect{z}_i^{(H)}\bigr]\\
        \vect{\tilde{y}_i} &= \text{LayerNorm}(\vect{x_i} + \vect{z}_i)
        \\
        \vect{y_i} &= \text{LayerNorm}(\vect{\tilde{y}_i} + \text{FC}(\text{ReLU}(\text{FC}(\vect{\tilde{y}_i})))
    \end{aligned}
\end{gather}

where FC is a fully-connected layer, LayerNorm is \emph{layer normalization},$1 \le h \le H$, $W_Q^{(h)}, W_K^{(h)}, W_V^{(h)} \in \mathbb{R}^{d_x \times (d_x / H)}$, and  $\vect{r_{ij}^K},\vect{r_{ij}^V}$ are learned embeddings for the relations between schema and question. We refer to \cite{wang2020} for full list of all relations.
To get a representation for the schema and question, a 
We use BART to encode the question and schema constants and apply $H=8$ layers of RAT layers.
% $q,Sq$
}

\paragraph{Autoregressive top-down decoding}
The prevailing method for decoding in semantic parsing has been grammar-based autoregressive top-down decoding \citep{yin2017syntactic,krishnamurthy2017neural,rabinovich2017abstract}, which guarantees decoding of syntactically valid programs. Specifically, the target program is represented as an abstract syntax tree under the grammar of the formal language, and linearized to a sequence of rules (or actions) using a top-down depth-first traversal. Once the program is represented as a sequence, it can be decoded using a standard sequence-to-sequence model with encoder attention \citep{dong2016}, often combined with beam search. We refer the reader to the aforementioned papers for further details on grammar-based decoding. 

We now turn to describe our method, which provides a radically different approach for decoding in semantic parsing.

\commentout{
A seq2seq encoder-decoder architecture is the standard for this task,
The gold query tree is linearized to the sequence of the preorder traversal.
The model predicts a single action at each decoding time-step, resulting in the top-down derivation of the logical form.
The loss is the NLL of the correct next derivation rule.
During Beam search, partial trees are used to determine the next actions, where the parts not yet predicted is in the bottom of the tree.
}
\section{The \ourparser{} parser}
\label{sec:model}

\begin{algorithm}[t]
\small
\textbf{input:} utterance $x$, schema $S$ \\
    $\bfx,\bfs \leftarrow \text{Encode}_{\text{RAT}}(x,S)$ \\
    %$Z_{0} \leftarrow \argmax_{K}(f_\text{const}(\bfs))$ \\ \label{line:init}
    $Z_{0} \leftarrow \text{Top-$K$ schema constants and DB values}$ \\ \label{line:init}
    \For{$t \leftarrow 0 \dots T-1$}{
        $Z'_{t} \leftarrow \text{Attention}(Z_t, \bfx, \bfx)$ \label{line:context}\\
        $F_{t+1} \leftarrow  \text{Score-frontier}(Z'_{t})$ \label{line:score}\\
        $F'_{t+1} \leftarrow \argmax_{K}(F_{t+1})$ \label{line:prune}\\
        $Z_{t+1} \leftarrow \text{Represent-beam}(Z_{t},F'_{t+1})$ \label{line:represent}\\
    %   $\mathcal{L}_{i}  \leftarrow CE(F_i,F_{i}\in  G_{i})$\\
    %   $Y_{i}  \leftarrow I_{i+1}\in  G$\\
    %   $\sum_{s \in G\cap I_{i+1}}{\log \sigma\left(x_{n}\right)}$\\
    %   $ l_{n}=y_{n} \cdot \log \sigma(x_{n})+(1-y_{n}) \cdot \log(1-\sigma(x_{n}))$
 }
 \Return $\argmax_z(Z_T)$
% $\text{Re-rank}(\bfx, \{Z_{t}\}^{T}_{t=0})$ %\label{line:rerank}
\caption{\ourparser{}}
\label{alg:highlevel}
\end{algorithm}

We first provide a high-level overview of \ourparser{} (see Algorithm~\ref{alg:highlevel} and Figure~\ref{fig:overview}).
As explained in \S\ref{sec:background}, we encode the utterance and schema with a RAT-SQL encoder. We initialize the beam (line~\ref{line:init}) with the $K$ highest scoring trees of height $0$, which include either schema constants or DB values. All trees are scored independently and in parallel, in a procedure formally defined in \S\ref{subsec:beaminit}.

Next, we start the search procedure. 
At every step $t$, attention is used to contextualize the trees with information from input question representation (line~\ref{line:context}). This representation is used to score every tree on the \emph{frontier}: the set of sub-trees of depth $\leq t+1$ that can be constructed from sub-trees on the beam with depth $\leq t$ (lines~\ref{line:score}-\ref{line:prune}). 
After choosing the top-$K$ trees for step $t+1$, we compute a new representation for them (line~\ref{line:represent}). Finally, we return the top-scoring tree from the final decoding step, $T$.
%we collect all trees constructed during the search procedure and re-rank them with another Transformer. 
Steps in our model operate on tree representations independently, and thus each step is efficiently parallelized.

\ourparser{} resembles beam search as in each step it holds the top-$K$ trees of a fixed height. It is also related to (pruned) chart parsing, since trees at step $t+1$ are computed from trees that were found at step $t$. This is unlike sequence-to-sequence models where items on the beam are competing hypotheses without any interaction.

\commentout{
On the first step we rank the RAT enriched schema items using a MLP, and initialize the first beam with the top-$K$ scoring items.
On step $t$, the beam contains trees of depth $t$, contextualize the beam with the question, and score every possible way to combine them to trees of depth $t+1$, we then take the top scoring $K$ candidates, and compute a representation for them.
After $k$ iterations, we apply another contextualization, and re-rank all the valid trees from all decoding steps, and return the top scoring tree.
}

%During search, at step $t$ we dynamically calculate the correct trees of depth $t+1$ and maximize the NLL of the probability of choosing them. 
%During re-ranking we maximize the MML of the correct trees.
%Similarly to \cite{guo2019complex} we use an intermediate representation (IR). Predictions of the model are made in IR and it is converted to SQL for evaluation.  

We now provide the details of our parser. First, we describe the formal language (\S\ref{subsec:rep}), then we provide precise details of our model architecture (\S\ref{subsec:architecture}) including beam initialization (\S\ref{subsec:beaminit}, we describe the training procedure (\S\ref{subsec:training}), and last, we discuss the properties of \ourparser{} compared to prior work (\S\ref{subsec:discussion}). 

\subsection{Representation of Query Trees}
\label{subsec:rep}
\paragraph{Relational algebra}
\newcite{guo2019complex} have shown recently that the mismatch between natural language and SQL leads to parsing difficulties. Therefore, they proposed SemQL, a formal query language with better alignment to natural language. 

\begin{table}[]
\scalebox{0.8}{
\begin{tabular}{llll}
Operation          &  Notation         & Input $\rightarrow$  Output   &  \\
\hline
Set Union             & $\cup$ &   $R \times R \rightarrow R$  &  \\
Set Intersection      & $\cap$ &  $R \times R \rightarrow R$ &  \\
Set difference    & $\setminus$ &  $R \times R\rightarrow R$  &  \\
Selection         & $\sigma$    &    $P \times R \rightarrow R$ &  \\
Cartesian product &$\times$    & $ R \times R \rightarrow R$ &  \\
Projection       &  $\Pi$         &  $C' \times R\rightarrow R$  &  \\
And                &$\land$         & $P \times P\rightarrow P$ &  \\
Or                & $\lor$             &  $P \times P\rightarrow P$ &  \\
Comparison                 & $\{\le\text{,}\ge\text{,}=\text{,}\neq\}$         &    $C \times C \rightarrow P$  &  \\
Constant Union          &     $\sqcup$         &  $C' \times C' \rightarrow C'$  & \\
Order by          &     $\tau_{asc/dsc}$         & $C \times R \rightarrow R$  &  \\
Group by          &     $\gamma$         & $C \times R \rightarrow R$  &  \\
Limit          &     $\lambda$         &  $C \times R \rightarrow R$  & \\
In/Not In          &     $\in,\not\in$         &  $C \times R \rightarrow P$  & \\
% Not              &  $\lnot$             &  $P \rightarrow P$  &  \\
Like/Not Like              &  $\sim,\not\sim$             &  $C \times C \rightarrow P$  &  \\
Aggregation  &   $\mathcal{G}_{\text{agg}}$ & $C \rightarrow C$  & \\
Distinct          &     $\delta$         &  $C \rightarrow C $  & \\
Keep          &     $\kappa$         & $\text{Any} \rightarrow \text{Any}$  &
\end{tabular}}
\caption{Our relational algebra grammar, along with the input and output semantic types of each operation. $P$: Predicate, $R$: Relation, $C$: schema  constant or DB value, $C'$: A set of constants/values, and $\text{agg} \in \{\text{sum}, \text{max}, \text{min}, \text{count}, \text{avg}\}$.}
\label{tab:grammar}
\end{table}

In this work, we follow their intuition, but instead of SemQL, we use the standard query language \emph{relational algebra} \citep{codd1970}. Relational algebra describes queries as trees, where leaves (terminals) are schema constants or DB values, and inner nodes (non-terminals) are operations (see Table~\ref{tab:grammar}). Similar to SemQL, its alignment with natural language is better than SQL. However, unlike SemQL, it is an existing query language, commonly used by SQL execution engines for query planning. 

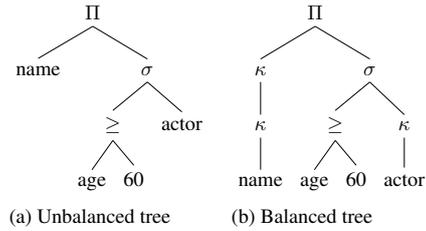
\begin{figure}
\centering
\scalebox{0.7}{
    \centering
    
    \begin{tabular}{ll}
\begin{tikzpicture} \Tree [.$\Pi$ name   [.$\sigma$  [.$\ge$ age  60  ] actor  ] ]\end{tikzpicture}
&
    %   \begin{tikzpicture}[level distance = 3em,sibling distance=5em]
        % \node[white] (start) {};
        
    %     \node[black] (project) {$\Pi$} {
    %       child {node [black] (keep1) {$\kappa$} child {node [black] (keep2) {$\kappa$} child {node [black] (name) {name}} }}
    %       child {node [black] (selection) {$\sigma$}
    %       child {node [black] (keep3) {$\kappa$} child {node [black] (actor) {actor}}}
    %       child {node [black] (ge) {$\ge$} child {node [black] (age) {age}} child {node [black] (sixty) {60}}}
    %       }
    %     };
        
    % \end{tikzpicture} \\[1ex]
    \begin{tikzpicture} \Tree [.$\Pi$  [.$\kappa$  [.$\kappa$ name  ] ]  [.$\sigma$  [.$\ge$ age  60  ]  [.$\kappa$ actor  ] ] ]\end{tikzpicture} \\
      (a) Unbalanced tree & (b) Balanced tree \\
    \end{tabular}
    }
    \caption{An unbalanced and balanced relational algebra tree (with the unary \textsc{Keep} operation) for the utterance \emph{``What are the names of actors older than 60?''}, where the corresponding SQL query is \texttt{SELECT name FROM actor WHERE age $\ge$ 60}.
    }
    \label{fig:keep}
    \vspace{-2ex}
\end{figure}

We write a grammar for relational algebra, augmented with SQL operators that are missing from relational algebra. We then
implement a transpiler that converts SQL queries to relational algebra for parsing, and then back from relational algebra to SQL for evaluation.
Table~\ref{tab:grammar} shows the full grammar, including the input and output semantic types of all operations. A relation ($R$) is a tuple (or tuples), a predicate ($P$) is a Boolean condition (evaluating to \texttt{True} or \texttt{False}), a constant ($C$) is a schema constant or DB value, and ($C'$) is a set of constants/values. Figure~\ref{fig:keep} shows an example relational algebra tree with the corresponding SQL query. More examples illustrating the correspondence between SQL and relational algebra (e.g., for the SQL \texttt{JOIN} operation) are in Appendix~\ref{app:relalg}. While our relational algebra grammar can also be adapted for standard top-down autoregressive parsing, we leave this endeavour for future work.

%Figure~\ref{fig:rel_algebra} shows an example SQL query and its translation to relation algebra, and Table~\ref{table:grammar} shows the grammar for relational algebra used by our parser to compose sub-trees. 
%Also note that our algorithm assumes that the grammar is in Chomsky Normal Form.

%We chose to use an IR based on a variant of Relational Algebra (RA), which is a formal language based on mathematical logic, and is the theoretical foundation for SQL. As a preprocessing step, we convert the train set into RA, our model outputs a prediction as RA, and is convert back into SQL for evaluation.
\paragraph{Tree balancing}
Conceptually, at each step \ourparser{} should generate new trees of height $\leq t+1$ and keep the top-$K$ trees computed so far. In practice, it is convenient to assume that trees are balanced. Thus, we want the beam at step $t$ to only have trees that are of height exactly $t$ (\emph{$t$-high trees}). 

To achieve this, we introduce a unary \textsc{Keep} operation that does not change the semantics of the sub-tree it is applied on. Hence, we can always grow the height of trees in the beam without changing the formal query.
For training (which we elaborate on in \S\ref{subsec:training}), we balance all relational algebra trees in the training set using the \textsc{keep} operation, such that the distance from the root to all leaves is equal. For example, in Figure~\ref{fig:keep}, two \textsc{Keep} operations are used to balance the column \texttt{actor.name}. After tree balancing, all constants and values are at height $0$, and the goal of the parser at step $t$ is to generate the gold set of $t$-high trees. 
% Details on the tree-balancing algorithm are in Appendix~\ref{app:balancing}.
 %A balanced tree is a tree where the distance from every leaf to the root is the same. For our algorithm to work, we need to work with balanced trees, so as a pre-processing step we pad the tree with special `keep' nodes. See Figure \ref{fig:keep}.

% \begin{figure}
%     \centering
%     \includegraphics[width=0.4\textwidth]{figs/keep.pdf}
%     \caption{Balancing the relational algebra tree with the unary \textsc{Keep} operation for the utterance \emph{"What are the names of actors older than 60?".}}
%     \label{fig:keep}
% \end{figure}

% \\

\subsection{Model Architecture}
\label{subsec:architecture}

\begin{figure}
    \centering
    \includegraphics[width=0.4\textwidth]{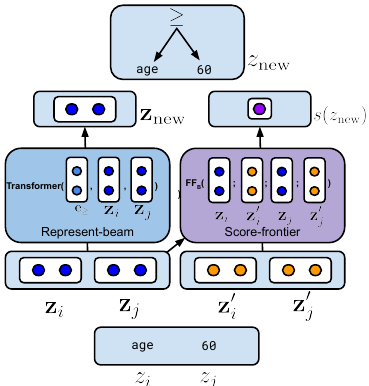}
    \caption{ Illustration of our tree scoring and representation mechanisms. $z$ is the symbolic tree, $\bfz$ is its vector representation, and $\bfz'$ its contextualized representation. }
    \label{fig:rep_score}
\end{figure}

To fully specify Alg.~\ref{alg:highlevel}, we need to define the following components: (a) scoring of trees on the frontier (lines~\ref{line:context}-\ref{line:score}), (b) representation of trees (line~\ref{line:represent}), and (c) representing and scoring of constants and DB values during beam initialization (leaves). We now describe these components. Figure~\ref{fig:rep_score} illustrates the scoring and representation of a binary operation. 

\paragraph{Scoring with contextualized beams} 
\ourparser{} maintains at each decoding step a beam $Z_t = ((z^{(t)}_1, \bfz^{(t)}_1), \dots, (z^{(t)}_K, \bfz^{(t)}_K))$, where $z^{(t)}_i$ is a symbolic representation of the query tree, and $\bfz^{(t)}_i$ is its corresponding vector representation. Unlike standard beam search, trees on our beams do not only compete with one another, but also \emph{compose} with each other (similar to chart parsing). For example, in Fig.~\ref{fig:overview}, the beam $Z_0$ contains the column \texttt{age} and the value \texttt{60}, which compose using the $\geq$ operator to form the $\texttt{age} \geq \texttt{60}$ tree.

We contextualize tree representations on the beam using cross-attention. Specifically, we use standard attention \cite{vaswani2017} to give tree representations access to the input question: $Z'_{t} \leftarrow \text{Attention}(Z_t, \bfx, \bfx)$, where the tree representations $(\bfz_{1}^{(t)}, \dots, \bfz_K^{(t)})$ 
are the queries, and the input tokens $(\bfx_1,\dots,\bfx_{|x|})$ are the keys and values.

Next, we compute scores for all $(t+1)$-high trees on the frontier. Trees can be generated by applying either a unary (including \textsc{Keep}) operation $u \in \sU$ or binary operation $b \in \sB$ on beam trees. Let $\bfw_u$ be a \emph{scoring vector} for a unary operation (such as $\bfw_\kappa$, $\bfw_\delta$, etc.), let $\bfw_b$ be a \emph{scoring vector} for a binary operation (such as $\bfw_\sigma$, $\bfw_\Pi$, etc.), and let $\textbf{z}'_i, \textbf{z}'_j$ be contextualized tree representations on the beam. 
We define a scoring function for frontier trees, where the score for a new tree $z_\text{new}$ generated by applying a unary rule $u$ on a tree $z_i$ is defined as follows:
$$s(z_\text{new}) = \bfw_u^\top FF_{U}([\textbf{z}_i; \textbf{z}_i']),$$ where $FF_U$ is a 2-hidden layer feed-forward layer with relu activations, and $[\cdot;\cdot]$ denotes concatenation.
Similarly the score for a tree generated by applying a binary rule $b$ on the trees $z_i, z_j$ is:
$$s(z_\text{new}) = \bfw_b^\top FF_{B}([\textbf{z}_i; \textbf{z}_i'; \textbf{z}_j; \textbf{z}_j']),$$ where $FF_B$ is another 2-hidden layer feed-forward layer with relu activations.

%We concatenate the contextualized and non-contextualized tree representation $\textbf{z}'' = [\textbf{z};\textbf{z}']$ and apply our scoring over this representation.
%We define a scoring function for frontier trees, where the score for a new tree $z_\text{new}$ generated by applying a unary rule $u$ on a tree $z_i$ is defined as following:
%$$s(z_\text{new}) = \bfw_u^\top \text{Relu}(W_{op}FF_{U}(\textbf{z}_{i}''))$$
%And similarly the score for a tree generated by applying a binary rule $b$ is
%$$s(z_\text{new}) = \bfw_b^\top \text{Relu}(W_{op}FF_{B}(W_{\ell}\textbf{z}_{i}''+W_{r}{\textbf{z}_{j}''}))$$

% $s(z_\text{new}) = \bfw_u^\top \textbf{z}'_i$, and similarly the score for a tree generated by applying a binary rule $b$ is
% $s(z_\text{new}) = 
% {\textbf{z}'}_{i}^{\top} W_b{\textbf{z}'}_{j}$. 

% Every item in the frontier consists of a label and one or two children.
%Where $W_{\ell}$ and $W_{r}$ are linear projections, $FF_{B}$ and $FF_{U}$ are $\text{Relu}$ feed-forward networks with a single hidden layer, and $W_{op}$ is a shared linear projection.

We use semantic types to detect invalid rule applications and fix their score to $s(z_\text{new}) = -\infty$. This guarantees that the trees \ourparser{} generates are well-formed, and the resulting SQL is executable.
\commentout{
We define the following simple linear scoring function for tree representations $\bfz_i,\bfz_j$ on the beam:
\begin{align*}
s(r) = \left\{\begin{matrix}
    \bfw_u^\top \bfz_i & \text{(unary)} \\ 
    \bfb_z^\top W_b\bfz_j & \text{(binary)} \\
\end{matrix}\right.\\    
\end{align*}
}
Overall, the total number of trees on the frontier is $\leq K|\sU| + K^2|\sB|$.
Because scores of different trees on the frontier are independent, they are efficiently computed in parallel. Note that we score new trees from the frontier \emph{before} creating a representation for them, which we describe next.

\paragraph{Recursive tree representation}
after scoring the frontier, we generate a recursive vector representation for the top-$K$ trees. While scoring is done with contextualized trees, representations are \emph{not} contextualized. We empirically found that contextualized tree representations slightly reduce performance, possibly due to optimization issues.

We represent trees with another standard Transformer layer. Let $\bfz_\text{new}$ be the representation for a new tree,
let $e_\ell$ be an embedding for a unary or binary operation, and let $\bfz_i, \bfz_j$   be non-contextualized tree representations from the beam we are extending. We compute a new representation as follows:
\begin{align*}
\bfz_\text{new} = \left\{\begin{matrix}
    \text{Transformer}(\bfe_\ell,\bfz_i) & \text{unary } \ell \\ 
    \text{Transformer}(\bfe_{\ell},\bfz_i,\bfz_j) & \text{binary } \ell \\
    \bfz_i & \ell = \textsc{Keep}
\end{matrix}\right.
\end{align*}
where for the unary \textsc{Keep} operation, we simply copy the representation from the previous step. 

\paragraph{Return value} As mentioned, the parser returns the highest-scoring tree in $Z_T$. More precisely, we return the highest-scoring \emph{returnable} tree, where a returnable tree is a tree that has a valid semantic type, that is, Relation (R).

\subsection{Beam initialization} 
\label{subsec:beaminit}
As described in Line~\ref{line:init} of Alg.~\ref{alg:highlevel}, the beam $Z_0$ is initialized with $K$ schema constants (e.g., \texttt{actor}, \texttt{age}) and DB values (e.g., \texttt{60}, \texttt{``France''}). In particular, we independently score schema constants and choose the top-$\frac{K}{2}$, and similarly score DB values and choose the top-$\frac{K}{2}$, resulting in a total beam of size $K$. 

\paragraph{Schema constants} We use a simple scoring function $f_\text{const}(\cdot)$. Recall that $\bfs_i$ is a representation of a constant, contextualized by the rest of the schema and the utterance. The function $f_\text{const}(\cdot)$ is a feed-forward network that scores each schema constant independently: $f_\text{const}(\bfs_i) =  \bfw_\text{const}\tanh{(W_{\text{const}}\bfs_i)}$, and the top-$\frac{K}{2}$ constants are placed in $Z_0$. 

\paragraph{DB values} Because the number of values in the DB is potentially huge, we do not score all DB values. Instead, we learn to detect spans in the question that correspond to DB values. This leads to a setup that is similar to extractive question answering \cite{rajpurkar-etal-2016-squad}, where the model outputs a distribution over input spans, and thus we adopt the architecture commonly used in extractive question answering. 
Concretely, we compute the probability that a token is the start token of a DB value, $P_\text{start}(x_i) \propto \exp(\bfw_\text{start}^\top \bfx_i)$, and similarly the probability that a token is the end token of a DB value, $P_\text{end}(x_i) \propto \exp(\bfw_\text{end}^\top \bfx_i)$, where $\bfw_\text{start}$ and $\bfw_\text{end}$ are parameter vectors. We define the probability of a span $(x_i, \dots, x_j)$ to be $P_\text{start}(x_i) \cdot P_\text{end}(x_j)$, and place in the beam $Z_0$ the top-$\frac{K}{2}$ input spans, where the representation of a span $(x_i, x_j)$ is the average of $\bfx_i$ and $\bfx_j$. 

A current limitation of \ourparser{} is that it cannot generate DB values that do not appear in the input question. This would require adding a mechanism such as \textsc{``bridge''} proposed by \newcite{lin-etal-2020-bridging}.

%Our neural representation of trees is recursive, Given a label $\ell$ and either one or two children $a_i,a_j$, we encode the tree with the root $\ell$ and children $a_i,a_j$ as the following: 
%For each binary/unary action we apply a TreeLSTM/LSTM \cite{tai2015improved} \cite{10.1162/neco.1997.9.8.1735} respectively with a learned action vector for that action as input.
%For the `keep` action, we don't modify the representation of the beam item. 
%$$a_{r} = \left\{\begin{matrix}
%    LSTM(e_{\ell},a_j) & r=(\ell\rightarrow a_j) \\ 
%    TreeLSTM(e_{\ell},a_j,a_k) & r=(\ell\rightarrow a_{j}a_{k}) \\
%    a_j & r=(keep\rightarrow a_{j})
%
%\end{matrix}\right.$$

% \end{tikzpicture}
\subsection{Training} \label{subsec:training}
To specify the loss function, we need to define the supervision signal. Recall that given the gold SQL program, we convert it into a gold \emph{balanced relational algebra tree} $z^\text{gold}$, as explained in \S\ref{subsec:rep} and Figure~\ref{fig:keep}. This lets us define for every decoding step the set of $t$-high gold sub-trees $\sZ_t^\text{gold}$. For example $\sZ_0^\text{gold}$ includes all gold schema constants and input spans that match a gold DB value,\footnote{In Spider, in 98.2\% of the training examples, all gold DB values appear as input spans.} $\sZ_1^\text{gold}$ includes all $1$-high gold trees, etc.

During training, we apply ``bottom-up Teacher Forcing'' \citep{williams1989}, that is, we populate\footnote{We compute this through an efficient tree hashing procedure. See Appendix \ref{app:hashing}. } the beam $Z_t$ with all trees from $\sZ_t^\text{gold}$ and then fill the rest of the beam (of size $K$) with the top-scoring non-gold predicted trees. This guarantees that we will be able to compute a loss at each decoding step, as described below.

%\ourparser{} is trained end-to-end, but it is not fully differentiable due to the $\argmax$ operation in line~\ref{line:prune} of Alg.~\ref{alg:highlevel}. Fortunately, we have gold trees that allow us to define a loss term at every decoding step, and we can train the parameters as usual with backpropagation.

\paragraph{Loss function}
During search, our goal is to give high scores to the possibly multiple \emph{sub-trees} of the gold tree. Because of teacher forcing, the frontier $F_{t+1}$ is guaranteed to contain all gold trees $\sZ_{t+1}^\text{gold}$. We first apply a softmax over all frontier trees
$p(z_\text{new}) = \text{softmax}\{s(z_\text{new})\}_{z_\text{new} \in F_{t+1}}$
and then maximize the probabilities of gold trees:
\begin{align*}
\frac{1}{C}\sum_{t=0}^{T}\sum_{z_t\in \sZ_{t}^\text{gold}}\log p\left(z_t\right)
\end{align*}
where the loss is normalized by $C$, the total number of summed terms. In the initial beam, $Z_0$, the probability of an input span is the product of the start and end probabilities, as explained in \S\ref{subsec:beaminit}.

\subsection{Discussion}
\label{subsec:discussion}

\begin{figure}
    \centering
    \includegraphics[width=0.5\textwidth]{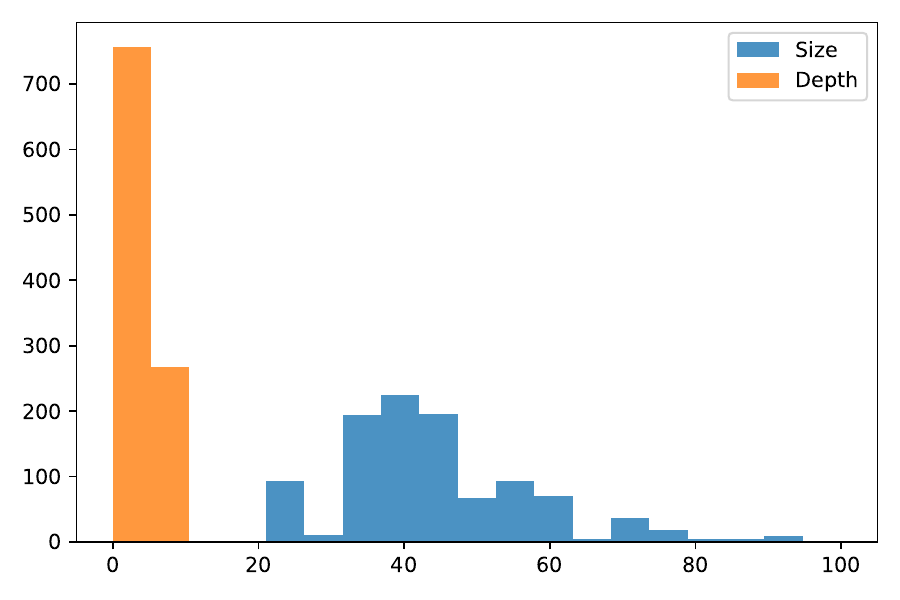}
    \caption{A histogram showing the distribution of
    the height of relational algebra trees in \textsc{Spider}, and the size of equivalent SQL query trees.}
    \label{fig:lenhist}
\end{figure}

To our knowledge, this work is the first to present a semi-autoregressive bottom-up semantic parser. We discuss the benefits of our approach.

\ourparser{} has theoretical runtime complexity that is logarithmic in the size of the tree instead of linear for autoregressive models. Figure~\ref{fig:lenhist} shows the distribution over the height of relational algebra trees in \textsc{Spider}, and the size of equivalent SQL query trees. Clearly, the height of most trees is at most 10, while the size is 30-50, illustrating the potential of our approach. In \S\ref{sec:experiments}, we demonstrate that indeed semi-autoregressive parsing leads to substantial empirical speed-up.

%In practice, achieving inference speedup requires fully parallelizing all decoding operations. Our current implementation does not support that and so inference is not faster than autoregressive models.

Unlike top-down autoregressive models, 
\ourparser{} naturally computes representations $\bfz$ for all sub-trees constructed at decoding time, which are well-defined semantic objects. These representations can be used in setups such as \emph{contextual semantic parsing}, where a semantic parser answers a sequence of questions. For example, given the questions \emph{``How many students are living in the dorms?''} and then \emph{``what are their last names?''}, the pronoun \emph{``their''} refers to a sub-tree from the SQL tree of the first question. Having a representation for such sub-trees can be useful when parsing the second question, in benchmarks such as \textsc{SPARC} \cite{Yu2019sparc}.

Another potential benefit of bottom-up parsing is that sub-queries can be executed while parsing \cite{berant2013freebase,liang2017nsm}, which can guide the search procedure. Recently, \newcite{odena2020bustle} proposed such an approach for program synthesis, and showed that conditioning on the results of execution can improve performance.  We do not explore this advantage of bottom-up parsing in this work, since executing queries at training time leads to a slow-down during training.

\ourparser{} is a bottom-up semi-autoregressive parser, but it could potentially be modified to be autoregressive by decoding one tree at a time. Past work \cite{Jianpeng2019} has shown that the performance of bottom-up and top-down autoregressive parsers is similar, but it is possible to re-examine this given recent advances in neural architectures.

\mycomment{
In this work, we score trees based on a contextualized representation, since trees do not only compete with one another, but also combine with each other. For example, a tree might get a higher score conditioned on another tree in the beam that it can compose with. Conceptually, beam contextualization can be done also  in autoregressive parsing, but the incentive is lower, as different items on the beam represent independent hypotheses.

Last, re-ranking has been repeatedly shown to be useful in semantic parsing \citep{goldman2017,yin2019}. However, in all prior work a re-ranker is a separate model, trained independently, that consumes the input examples from scratch. In \ourparser{}, we compute tree representations that can be ranked inside the model with a single Transformer-based re-ranking layer.
}
\section{Experimental Evaluation}
\label{sec:experiments}

We conduct our experimental evaluation on \textsc{Spider} \citep{yu2019}, a challenging large-scale dataset for text-to-SQL parsing.
\textsc{Spider} has become a common benchmark for evaluating semantic parsers because it includes complex SQL queries and a realistic zero-shot setup, where schemas at test time are different from training time.

\subsection{Experimental setup}
We encode the input utterance $x$ and the schema $S$ with \textsc{Grappa}, consisting of 24 Transformer layers, followed by another 8 RAT-SQL layers, which we implement inside AllenNLP \cite{gardner2018allennlp}. Our beam size is $K=30$, and the number of decoding steps is $T=9$ at inference time, which is the maximal tree depth on the development set.
%\jh{This seems limiting, as in reality trees are not bounded. I'm wondering whether using larger T, just in case, affects the performance}
The transformer used for tree representations has one layer, 8 heads, and dimensionality 256. We train for 60K steps with batch size 60, and perform early stopping based on the development set.
%The official evaluation is over \emph{anonymized SQL queries}, that is all DB values (that are \emph{not} schema items) are anonymized to \texttt{value}, and thus we train and test our model with anonymized SQL queries (see more about using DB content below).

\paragraph{Evaluation}
We evaluate performance with the official \textsc{Spider} evaluation script, which computes \emph{exact match (EM)}, i.e., whether the predicted SQL query is identical to the gold query after some query normalization. The evaluation script uses \emph{anonymized} queries, where DB values are converted to a special \texttt{value} token. In addition, for models that output DB values, the evaluation script computes \emph{denotation accuracy}, that is, whether executing the output SQL query results in the right denotation (answer). As \ourparser{} generates DB values, we evaluate using both EM and denotation accuracy

%We cannot compare the runtime of \ourparser{} to  autoregressive decoder \textsc{RAT-SQL+Grappa} is impossible, since code is not available. Instead we use the decoder from  \textsc{RAT-SQLv3+BERT}, which is identical to the best of our knowledge. This autoregressive decoder uses beam search of size $1$ for the first few decoding steps, and then transitions to beam size $6$. 

\paragraph{Models}
We compare \ourparser{} to the best non-anonymous models on the \textsc{Spider} leaderboard at the time of writing. Our model is most comparable to \textsc{RAT-SQL+Grappa}, which has the same encoder, but an autoregressive decoder. 

In addition, we perform the following ablations and oracle experiments:
\begin{itemize}[leftmargin=*,itemsep=0pt,topsep=0pt]
    \item \textsc{No X-Attention}: We remove the cross attention that computes $Z'_t$ and uses the representations in $Z_t$ directly to score the frontier. In this setup, the decoder only observes the input question through the $0$-high trees in $Z_0$.
    \item \textsc{With Cntx Rep.}: We use the contextualized representations not only for \emph{scoring}, but also as input for creating the new trees $Z_{t+1}$. This tests if contextualized representations on the beam hurt or improve performance.
    \item \textsc{No DB Values}: We anonymize all SQL queries by replacing DB values with \texttt{value}, as described above, and evaluate \ourparser{} using EM. This tests whether learning from DB values improves performance.
    \item \textsc{$Z_{0}$-Oracle}: An oracle experiment where $Z_0$ is populated with the gold schema constants (but predicted DB values). This shows results given perfect schema matching.
\end{itemize}

\begin{table}[t]
\centering
\scalebox{1.0}{
{\footnotesize
\begin{tabular}{|l l l |}
\hline
Model                      & EM   &  Exec \\ \hline
\textsc{RAT-SQL}+\textsc{GP}+\textsc{Grappa}  & \textbf{69.8}\% &  n/a    \\ 
\textsc{RAT-SQL}+\textsc{GAP}  & 69.7\% &  n/a    \\ 
\textsc{RAT-SQL}+\textsc{Grappa}       &  69.6\%     &  n/a  \\ 
\textsc{RAT-SQL}+\textsc{STRUG}       &  68.4\%     &  n/a  \\ 
\textsc{BRIDGE}+\textsc{BERT} (ensemble)  & 67.5\% &  68.3    \\ 
\textsc{RAT-SQL}v3+\textsc{BERT}  & 65.6\% &  n/a    \\ 
% \textsc{RYANSQL}v2+\textsc{BERT}      & 60.6\%     \\ 
\hline
\textsc{\ourparser{}+\textsc{Grappa}}                     &  69.5\%     &  \textbf{71.1}\%  \\
% \textsc{OurRATSQL-large }              & XX.X\% & n/a \\
% \textsc{OurRATSQL-base }              & 65.3\% & n/a \\
\hline
\end{tabular}
}

}
\caption{Results on the \textsc{Spider} test set.}
\label{tab:results}
\end{table}

\begin{figure}[t]
    \centering
    \includegraphics[width=0.45\textwidth]{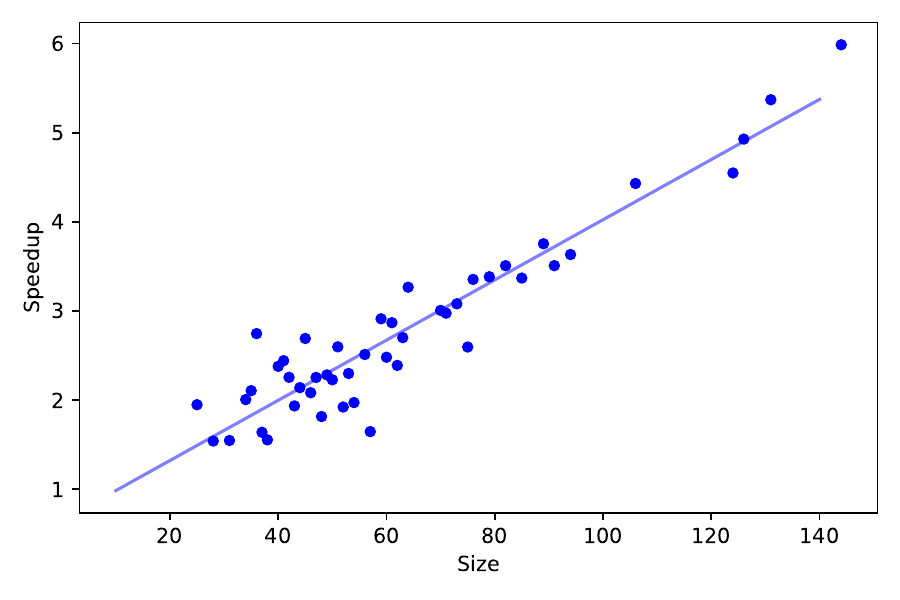}
    \caption{Speed-up on the development set compared to autoregressive decoding, w.r.t the size of the SQL query.}
    \label{fig:speedup}
\end{figure}
\begin{figure}[t]
    \centering
     \includegraphics[width=0.45\textwidth]{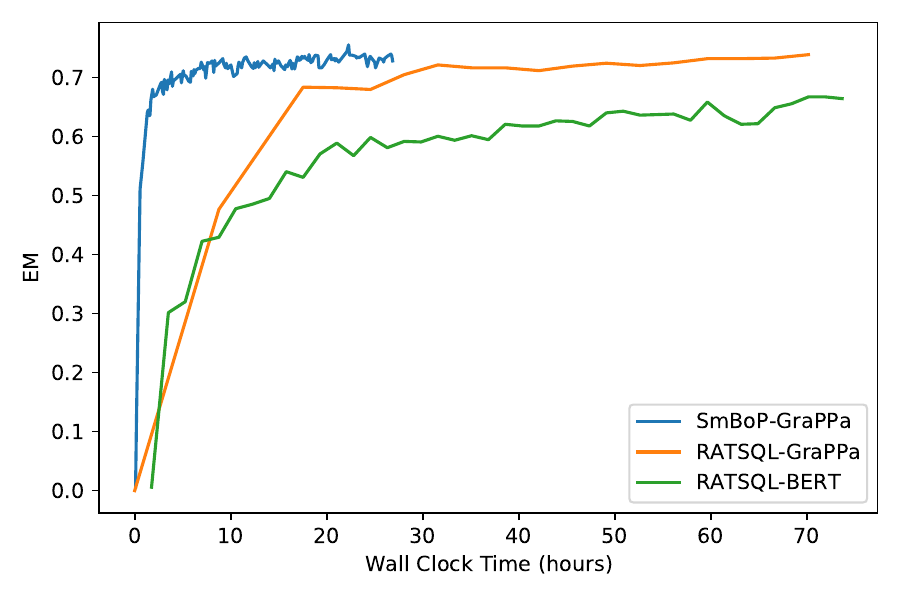}
    \caption{EM as a function of wall clock time on the development set of \textsc{SPIDER} during training.
}
    \label{fig:clock}
\end{figure}
\begin{figure}[t]
    \centering
     \includegraphics[width=0.45\textwidth]{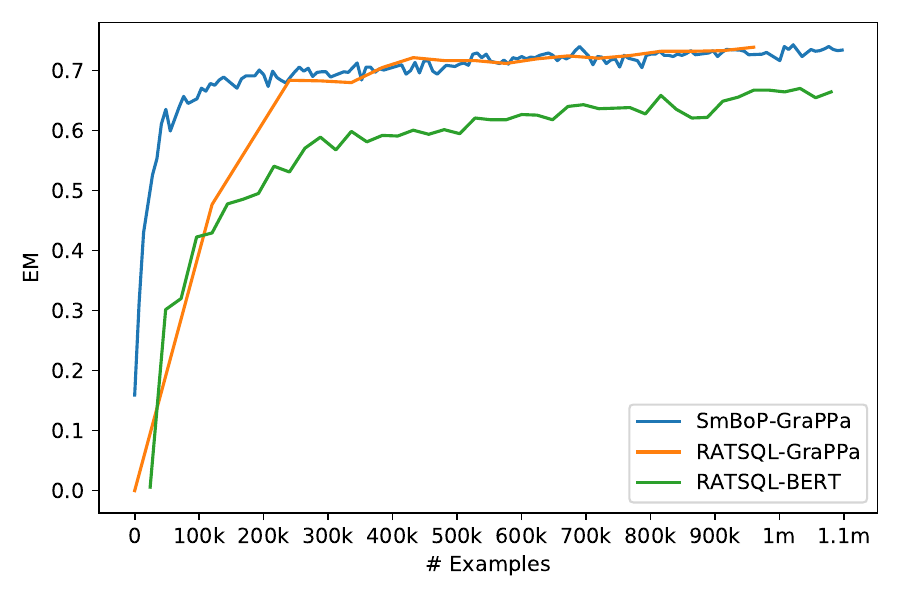}
    \caption{EM as a function of the number of examples on the development set of \textsc{SPIDER} during training.}
    \label{fig:conv_speed}
\end{figure}

% \begin{table}[t]
% \centering
% \scalebox{0.9}{
% {\footnotesize
% \begin{tabular}{|l l l  l|}
% \hline
% Model                     & EM & \textsc{BEM} &  \textsc{$Z_{0}$ recall}  \\ \hline
% \textsc{RAT-SQL+Grappa}              & 73.9\% & n/a & n/a            \\
% \hline
% \textsc{\ourparser{}}                    & 70.6\%  &  81.8\%  &    97.3\%  \\ 
% - \textsc{No Beam Cntx}                  & 71.7\%  &  81.9\%  & 98.3\% \\
% - \textsc{With cntx Rep.}         & 69.6\%  & 80.5\% &      96.7\%  \\ 
% \hline
% \textsc{\ourparser{}-$Z_{0}$-Oracle}     & 79.3\% & 88.5\% & n/a     \\ 
% \hline
% \end{tabular}
% }

% }
% \caption{Development set EM, beam EM (BEM) and recall on DB constants ($Z_0$ recall) for all models.
% }
% \label{tab:ablations_old}
% \end{table}

\begin{table}[t]
\centering
\scalebox{0.9}{
{\footnotesize
\begin{tabular}{|l l l l l|}
\hline
Model                     & Exec & EM & \textsc{BEM} &  \textsc{$Z_{0}$ rec.}  \\ \hline

\textsc{RAT-SQL}+\textsc{Grappa}             & n/a& 73.9\% & n/a & n/a            \\
\hline
\textsc{\ourparser{}}                    & 75.0\%& 74.7\%  &  82.6\%  &    98.3\%  \\ 
- \textsc{No X-Att}                  & 74.8\%& 72.7\%  &  81.1\%  & 96.6\% \\
- \textsc{With Cntx Rep.}        & 73.4\%& 72.4\%  & 81.9\% &      97.3\%  \\ 
- \textsc{No DB Values}         & n/a & 71.3\%  & 80.0\% &      98.1\%  \\ 
% - \textsc{Without DB Values}         & 70.1\%  & 80.6\% &      97.4\%  \\
\hline
\textsc{\ourparser{}-$Z_{0}$-Oracle}     & 77.4\% & 79.1\% & 85.8\% & n/a     \\ 
\hline
\end{tabular}
}
}
\caption{Development set EM, beam EM (BEM) and recall on schema constants and DB values ($Z_0$ rec.) for all models.
%\jh{I'm wondering how the same model with an independent reranker would perform. I guess it is a lot of work, but you claim the training everything together is an advantage} \oh{we wanted to do this, maybe for NAACL}
}
\label{tab:ablations}
\end{table}

\subsection{Results}

Table~\ref{tab:results} shows test results of \ourparser{} compared to the top 
(non-anonymous) entries on the leaderboard \cite{zhao2021gp,shi2021learning,yu2020grappa,deng2020structure,lin-etal-2020-bridging,rat-sql}. \ourparser{} obtains an EM of 69.5\%, only 0.3\% lower than the best model, and 0.1\% lower than \textsc{RAT-SQL}+\textsc{Grappa}, which has the same encoder, but an autoregressive decoder. Moreover, \ourparser{} outputs DB values, unlike other models that output anonymized queries that cannot be executed. \ourparser{} establishes a new state-of-the-art in denotation accuracy, surpassing an ensemble of \textsc{BRIDGE}+\textsc{BERT} models by 2.9 denotation accuracy points, and 2 EM points.

Turning to decoding time, we compare \ourparser{} to \textsc{RAT-SQL}v3+\textsc{BERT}, since 
the code for \textsc{RAT-SQL}v3+\textsc{Grappa} was not available. To the best of our knowledge the decoder in both is identical, so this should not affect decoding time. We find
that the decoder of \ourparser{} is on average 2.23x faster than the autoregressive decoder on the development set. Figure~\ref{fig:speedup} shows the average speed-up for different query tree sizes, where we observe a clear linear speed-up as a function of query size. For long queries the speed-up factor reaches 4x-6x. When including also the encoder, the average speed-up obtained by \ourparser{} is 1.55x. 

In terms of training time, \ourparser{} leads to much faster training and convergence. We compare the learning curves of \ourparser{} and \textsc{RAT-SQL}v3+\textsc{BERT}, both trained on an RTX 3090, and also to \textsc{RAT-SQL}v3+\textsc{Grappa} using performance as a function of the number of examples, sent to us in a personal communication from the authors. \ourparser{} converges much faster than \textsc{RAT-SQL} (Fig.~\ref{fig:conv_speed}). E.g., after 120K examples, the EM of \ourparser{} is 67.5,
while for \textsc{RAT-SQL+Grappa} it is 47.6. Moreover, \ourparser{} processes at training time 20.4 examples per second, compared to only 3.8 for the official \textsc{RAT-SQL} implementation. Combining these two facts leads to much faster training time (Fig.~\ref{fig:clock}), slighly more than one day for \ourparser{} vs. 5-6 days for \textsc{RAT-SQL}.

\paragraph{Ablations} Table~\ref{tab:ablations} shows results of ablations on the development set.
Apart from EM, we also report: (a) beam EM (BEM): whether a correct tree was found \emph{anywhere} during the $T$ decoding steps, and (b) \emph{$Z_0$ recall}: the fraction of examples where the parser placed all gold schema constants and DB values in $Z_0$. This estimates the ability of our models to perform schema matching in a single non-autoregressive step.

We observe that ablating cross-attention leads to a small reduction in EM. This rather small drop is surprising since it means that all information about the question is passed to the decoder through $Z_0$. We hypothesize that this is possible, because the number of decoding steps is
small, and thus utterance information can propagate through the decoder. Using contextualized representations for trees also leads to a small drop in performance. Last, we see that feeding the model with actual DB values rather than an anonymized \texttt{value} token improves performance by 3.4 EM points.

Looking at \textsc{$Z_{0}$ recall}, we see that
models perform well at detecting relevant schema constants and DB values (96.6\%-98.3\%), 
despite the fact that this step is fully non-autoregressive. However, an oracle model that places all gold schema constants and only gold schema constants in $Z_0$ further improves EM (74.7   $\rightarrow$79.1\%), with a BEM of 85.8\%. This shows that better schema matching and search can still improve performance on \textsc{Spider}.

BEM is 8\%-9\% higher than EM, showing that, similar to past findings in semantic parsing \cite{goldman2017,yin2019}, adding a re-ranker on top of the trees computed by \ourparser{} can potentially improve performance. We leave this for future work.

\subsection{Analysis}

\commentout{
\begin{figure}[t]
    \centering
    \includegraphics[width=0.45\textwidth]{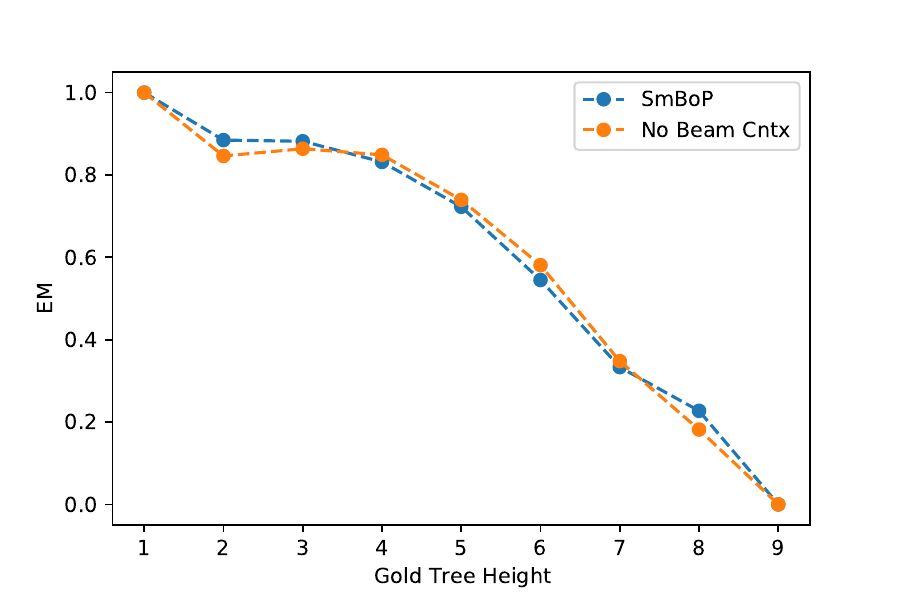}
    \caption{Breakdown of EM across different gold tree heights. We observe contextualizing trees improves performance specifically for examples with deep trees.}
    \label{fig:depth}
\end{figure}
}
%Figure~\ref{fig:depth} breaks down EM performance based on the height of the gold tree. Similar to auto-regressive models, performance drops as the height increases. Interestingly, the beam and re-ranking Transformers substantially improve performance for heights 7-8, showing the benefit of contextualizing trees in the more difficult examples. All models fails on the 15 examples with trees of height $>8$.

\commentout{
\begin{figure}
    \centering
    \includegraphics[width=0.5\textwidth]{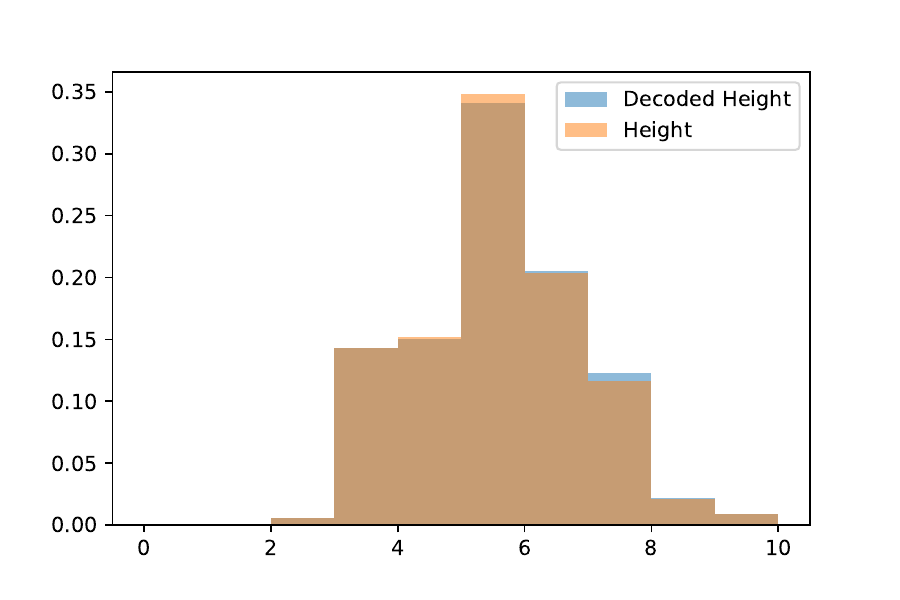}
    \caption{Distribution of heights for gold and predicted trees.}
    \label{fig:dec_height}
\end{figure}
}

\begin{figure}[t]
    \centering
    \includegraphics[width=0.4\textwidth]{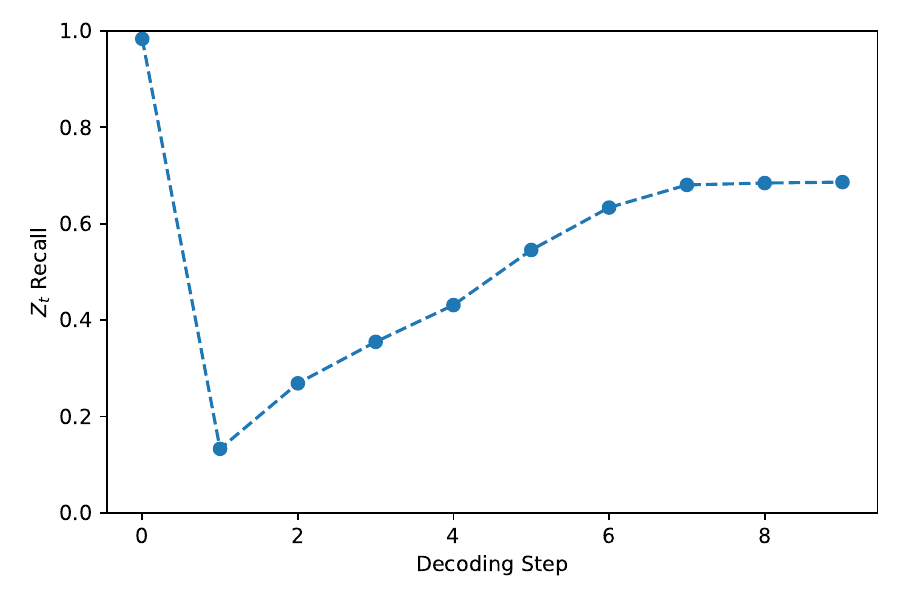}
    \caption{$Z_{t}$ Recall across decoding steps.}
    \label{fig:regret}
\end{figure}

We extend the notion of $Z_{0}$ recall to all decoding steps, where $Z_{t}$ recall is whether all gold $t$-high sub-trees were generated at step $t$. We see $Z_{t}$ recall across decoding steps in Figure~\ref{fig:regret}.\footnote{This metric checks for exact sub-tree match, unlike EM that does more normalization, so numbers are not comparable to EM.} The drop after step $0$ and subsequent rise indicate that the model maintains in the beam, using the \textsc{Keep} operation, trees that are sub-trees of the gold tree, and expands them in later steps. This means that the parser can recover from errors in early decoding steps as long as the relevant trees are kept on the beam.

To better understand search errors we perform the following analysis. For each example, we find the first gold tree that is dropped from the beam (if there is more than one, we choose one randomly). We then look at the children of $t$, and see whether at least one was expanded in some later step in decoding, or whether the children were completely abandoned by the search procedure. We find that in 62\% of the cases indeed one of the children was incorrectly expanded, indicating a composition error.

\commentout{
In fig \ref{fig:dec_height} we can we the distribution of the depth for the gold trees vs the distribution of the depth of the decoded trees, for the correct examples. Looking at the graph, we can see that during search we are able to find the gold tree with minimal number of extra steps, this is impressive since the model has the ability to regret past decisions, which it does not use. \oh{this is done via comparing hash values, I need to change this to EM, maybe it will regret stuff there}
}

\commentout{
\begin{figure}[t]
    \centering
    \includegraphics[width=0.4\textwidth]{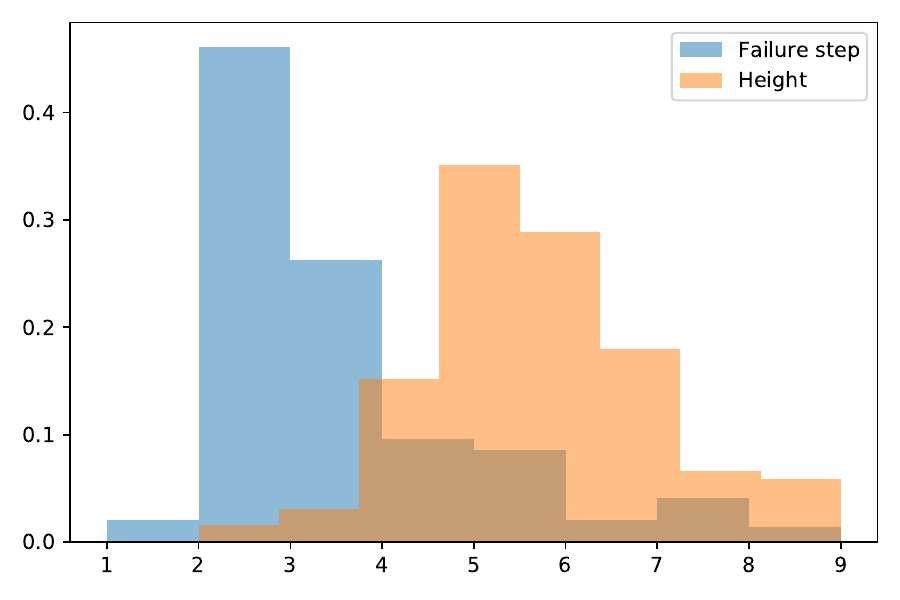}
    \caption{Distribution of the failure step and gold height for incorrect examples.}
    \label{fig:failure_step}
\end{figure}

\ourparser{} fails if at some decoding step, it does not add to the beam a sub-tree necessary for the final gold tree, and no other beam trees can be combined to create it. To measure the performance of our search procedure, the histogram in Figure~\ref{fig:failure_step} shows the distribution over steps where failure occurs, as well as the distribution over tree heights of errors. We observe that when our model is wrong, most of the errors are early in the search procedure, possibly because more gold trees need to be produced in each step.
}

%At the start of decoding, the model makes less mistakes, but once the beam fills with relevant items, it becomes harder to select the correct sub-trees. once we get into decoding step 6, the model makes much less mistakes, since there are less options to combine two items.

In this work, we used beam size $K=30$. Reducing $K$ to $20$ leads to a drop of less than point (74.7$\rightarrow$73.8), and increasing $K$ to $40$ reduces performance by (74.7$\rightarrow$72.6). In all cases, decoding time does not dramatically change.

%We compare multiple values of the beam size $K$ and see that a smaller $K$  does not lead to a substantial boost in examples per second. We also note that increasing the beam size does not improve EM, and in fact causes instability during training.

Last, we randomly sample 50 errors from \ourparser{} and categorize them into the following types:
\begin{itemize}[leftmargin=*,itemsep=0pt,topsep=0pt]
    \item Search errors (52\%): we find that most search errors are due to either extra or missing \texttt{JOIN} or \texttt{WHERE} conditions .
    \item Schema encoding errors (34\%): Missing or extra schema constants in the predicted query.
    \item  Equivalent queries (12\%): Predicted trees that are equivalent to the gold tree, but the automatic evaluation script does not handle.
\end{itemize}

\section{Conclusions}
In this work we present the first semi-autoregressive bottom-up semantic parser
that enjoys logarithmic theoretical runtime, and show that it leads to a 2.2x
speed-up in decoding and $\sim$5x faster taining, while maintaining state-of-the-art performance.
Our work shows that bottom-up parsing, where the model learns representations for semantically meaningful sub-trees is a promising research direction, that can contribute in the future to setups such as contextual semantic parsing, where sub-trees often repeat, and can enjoy the benefits of execution at training time. Future work can also leverage work on learning tree representations  \cite{Shiv2018NovelPE} to further improve parser performance.

\section*{Acknowledgments}
We thank Tao Yu, Ben Bogin, Jonathan Herzig, Inbar Oren, Elad Segal and Ankit Gupta for their useful comments. This research was partially supported by The Yandex Initiative for Machine Learning, and the European Research Council (ERC) under the European
Union Horizons 2020 research and innovation programme (grant ERC DELPHI 802800).

\bibliographystyle{acl_natbib}
\bibliography{naacl2021}

\begin{thebibliography}{37}
\expandafter\ifx\csname natexlab\endcsname\relax\def\natexlab#1{#1}\fi

\bibitem[{Berant et~al.(2013)Berant, Chou, Frostig, and
  Liang}]{berant2013freebase}
J.~Berant, A.~Chou, R.~Frostig, and P.~Liang. 2013.
\newblock Semantic parsing on {F}reebase from question-answer pairs.
\newblock In \emph{Empirical Methods in Natural Language Processing (EMNLP)}.

\bibitem[{Cheng et~al.(2019)Cheng, Reddy, Saraswat, and Lapata}]{Jianpeng2019}
Jianpeng Cheng, Siva Reddy, Vijay Saraswat, and Mirella Lapata. 2019.
\newblock \href {https://doi.org/10.1162/coli\_a\_00342} {Learning an
  executable neural semantic parser}.
\newblock \emph{Computational Linguistics}, 45(1):59--94.

\bibitem[{Clarke et~al.()Clarke, Goldwasser, Chang, and Roth}]{clarke2010}
James Clarke, Dan Goldwasser, Ming-Wei Chang, and Dan Roth.
\newblock \href {https://www.aclweb.org/anthology/W10-2903} {Driving semantic
  parsing from the world{'}s response}.
\newblock In \emph{Proceedings of the Fourteenth Conference on Computational
  Natural Language Learning (CoNLL)}.

\bibitem[{Codd(1970)}]{codd1970}
E.~F. Codd. 1970.
\newblock \href {https://doi.org/10.1145/362384.362685} {A relational model of
  data for large shared data banks}.
\newblock \emph{Commun. ACM}, 13(6):377–387.

\bibitem[{Deng et~al.(2020)Deng, Awadallah, Meek, Polozov, Sun, and
  Richardson}]{deng2020structure}
Xiang Deng, Ahmed~Hassan Awadallah, Christopher Meek, Oleksandr Polozov, Huan
  Sun, and Matthew Richardson. 2020.
\newblock Structure-grounded pretraining for text-to-sql.
\newblock \emph{arXiv preprint arXiv:2010.12773}.

\bibitem[{Devlin et~al.(2019)Devlin, Chang, Lee, and
  Toutanova}]{devlin-etal-2019-bert}
Jacob Devlin, Ming-Wei Chang, Kenton Lee, and Kristina Toutanova. 2019.
\newblock \href {https://doi.org/10.18653/v1/N19-1423} {{BERT}: Pre-training of
  deep bidirectional transformers for language understanding}.
\newblock In \emph{Proceedings of the 2019 Conference of the North {A}merican
  Chapter of the Association for Computational Linguistics: Human Language
  Technologies, Volume 1 (Long and Short Papers)}, pages 4171--4186,
  Minneapolis, Minnesota. Association for Computational Linguistics.

\bibitem[{Dong and Lapata(2016)}]{dong2016}
Li~Dong and Mirella Lapata. 2016.
\newblock \href {https://doi.org/10.18653/v1/P16-1004} {Language to logical
  form with neural attention}.
\newblock In \emph{Proceedings of the 54th Annual Meeting of the Association
  for Computational Linguistics (ACL)}, pages 33--43, Berlin, Germany.
  Association for Computational Linguistics.

\bibitem[{Gardner et~al.(2018)Gardner, Grus, Neumann, Tafjord, Dasigi, Liu,
  Peters, Schmitz, and Zettlemoyer}]{gardner2018allennlp}
Matt Gardner, Joel Grus, Mark Neumann, Oyvind Tafjord, Pradeep Dasigi,
  Nelson~F. Liu, Matthew Peters, Michael Schmitz, and Luke Zettlemoyer. 2018.
\newblock \href {https://doi.org/10.18653/v1/W18-2501} {{A}llen{NLP}: A deep
  semantic natural language processing platform}.
\newblock In \emph{Proceedings of Workshop for {NLP} Open Source Software
  ({NLP}-{OSS})}, pages 1--6, Melbourne, Australia. Association for
  Computational Linguistics.

\bibitem[{Goldman et~al.(2018)Goldman, Latcinnik, Nave, Globerson, and
  Berant}]{goldman2017}
Omer Goldman, Veronica Latcinnik, Ehud Nave, Amir Globerson, and Jonathan
  Berant. 2018.
\newblock \href {https://doi.org/10.18653/v1/P18-1168} {Weakly supervised
  semantic parsing with abstract examples}.
\newblock In \emph{Proceedings of the 56th Annual Meeting of the Association
  for Computational Linguistics (ACL) (Volume 1: Long Papers)}, pages
  1809--1819, Melbourne, Australia. Association for Computational Linguistics.

\bibitem[{Guo et~al.(2019)Guo, Zhan, Gao, Xiao, Lou, Liu, and
  Zhang}]{guo2019complex}
Jiaqi Guo, Zecheng Zhan, Yan Gao, Yan Xiao, Jian-Guang Lou, Ting Liu, and
  Dongmei Zhang. 2019.
\newblock \href {https://doi.org/10.18653/v1/P19-1444} {Towards complex
  text-to-{SQL} in cross-domain database with intermediate representation}.
\newblock In \emph{Proceedings of the 57th Annual Meeting of the Association
  for Computational Linguistics (ACL)}, pages 4524--4535, Florence, Italy.
  Association for Computational Linguistics.

\bibitem[{Herzig et~al.(2020)Herzig, Nowak, M{\"u}ller, Piccinno, and
  Eisenschlos}]{herzig-etal-2020-tapas}
Jonathan Herzig, Pawel~Krzysztof Nowak, Thomas M{\"u}ller, Francesco Piccinno,
  and Julian Eisenschlos. 2020.
\newblock \href {https://doi.org/10.18653/v1/2020.acl-main.398} {{T}a{P}as:
  Weakly supervised table parsing via pre-training}.
\newblock In \emph{Proceedings of the 58th Annual Meeting of the Association
  for Computational Linguistics}, pages 4320--4333, Online. Association for
  Computational Linguistics.

\bibitem[{Krishnamurthy et~al.(2017)Krishnamurthy, Dasigi, and
  Gardner}]{krishnamurthy2017neural}
Jayant Krishnamurthy, Pradeep Dasigi, and Matt Gardner. 2017.
\newblock Neural semantic parsing with type constraints for semi-structured
  tables.
\newblock In \emph{Proceedings of the Conference on Empirical Methods in
  Natural Language Processing (EMNLP)}.

\bibitem[{Liang et~al.(2017)Liang, Berant, Le, and Lao}]{liang2017nsm}
C.~Liang, J.~Berant, Q.~Le, and K.~D. F.~N. Lao. 2017.
\newblock Neural symbolic machines: Learning semantic parsers on {F}reebase
  with weak supervision.
\newblock In \emph{Association for Computational Linguistics (ACL)}.

\bibitem[{Liang et~al.(2011)Liang, Jordan, and Klein}]{liang2011}
Percy Liang, Michael Jordan, and Dan Klein. 2011.
\newblock \href {https://www.aclweb.org/anthology/P11-1060} {Learning
  dependency-based compositional semantics}.
\newblock In \emph{Proceedings of the 49th Annual Meeting of the Association
  for Computational Linguistics: Human Language Technologies (ACL-HLT)}, pages
  590--599, Portland, Oregon, USA. Association for Computational Linguistics.

\bibitem[{Lin et~al.(2020)Lin, Socher, and Xiong}]{lin-etal-2020-bridging}
Xi~Victoria Lin, Richard Socher, and Caiming Xiong. 2020.
\newblock \href {https://doi.org/10.18653/v1/2020.findings-emnlp.438} {Bridging
  textual and tabular data for cross-domain text-to-{SQL} semantic parsing}.
\newblock In \emph{Findings of the Association for Computational Linguistics:
  EMNLP 2020}, pages 4870--4888, Online. Association for Computational
  Linguistics.

\bibitem[{Liu et~al.(2019)Liu, Ott, Goyal, Du, Joshi, Chen, Levy, Lewis,
  Zettlemoyer, and Stoyanov}]{liu2019roberta}
Yinhan Liu, Myle Ott, Naman Goyal, Jingfei Du, Mandar Joshi, Danqi Chen, Omer
  Levy, Mike Lewis, Luke Zettlemoyer, and Veselin Stoyanov. 2019.
\newblock Roberta: A robustly optimized bert pretraining approach.
\newblock \emph{arXiv preprint arXiv:1907.11692}.

\bibitem[{Merkle(1987)}]{DBLP:conf/crypto/Merkle87}
Ralph~C. Merkle. 1987.
\newblock \href {https://doi.org/10.1007/3-540-48184-2\_32} {A digital
  signature based on a conventional encryption function}.
\newblock In \emph{Advances in Cryptology - {CRYPTO} '87, {A} Conference on the
  Theory and Applications of Cryptographic Techniques, Santa Barbara,
  California, USA, August 16-20, 1987, Proceedings}, volume 293 of
  \emph{Lecture Notes in Computer Science}.

\bibitem[{Odena et~al.(2020)Odena, Shi, Bieber, Singh, and
  Sutton}]{odena2020bustle}
Augustus Odena, Kensen Shi, David Bieber, Rishabh Singh, and Charles Sutton.
  2020.
\newblock \href {http://arxiv.org/abs/2007.14381} {Bustle: Bottom-up
  program-synthesis through learning-guided exploration}.

\bibitem[{Rabinovich et~al.(2017)Rabinovich, Stern, and
  Klein}]{rabinovich2017abstract}
Maxim Rabinovich, Mitchell Stern, and Dan Klein. 2017.
\newblock \href {https://doi.org/10.18653/v1/P17-1105} {Abstract syntax
  networks for code generation and semantic parsing}.
\newblock In \emph{Proceedings of the 55th Annual Meeting of the Association
  for Computational Linguistics (ACL)}, pages 1139--1149, Vancouver, Canada.
  Association for Computational Linguistics.

\bibitem[{Rajpurkar et~al.(2016)Rajpurkar, Zhang, Lopyrev, and
  Liang}]{rajpurkar-etal-2016-squad}
Pranav Rajpurkar, Jian Zhang, Konstantin Lopyrev, and Percy Liang. 2016.
\newblock \href {https://doi.org/10.18653/v1/D16-1264} {{SQ}u{AD}: 100,000+
  questions for machine comprehension of text}.
\newblock In \emph{Proceedings of the 2016 Conference on Empirical Methods in
  Natural Language Processing}, pages 2383--2392, Austin, Texas. Association
  for Computational Linguistics.

\bibitem[{Schwartz et~al.(2020)Schwartz, Dodge, Smith, and
  Etzioni}]{Schwartz2020green}
Roy Schwartz, Jesse Dodge, Noah~A. Smith, and Oren Etzioni. 2020.
\newblock Green {AI}.
\newblock \emph{Communications of the ACM}, 63.

\bibitem[{Shaw et~al.(2018)Shaw, Uszkoreit, and Vaswani}]{shaw2018}
Peter Shaw, Jakob Uszkoreit, and Ashish Vaswani. 2018.
\newblock \href {https://doi.org/10.18653/v1/N18-2074} {Self-attention with
  relative position representations}.
\newblock In \emph{Proceedings of the 2018 Conference of the North {A}merican
  Chapter of the Association for Computational Linguistics: Human Language
  Technologies (NAACL-HLT), Volume 2 (Short Papers)}, pages 464--468, New
  Orleans, Louisiana. Association for Computational Linguistics.

\bibitem[{Shi et~al.(2021)Shi, Ng, Wang, Zhu, Li, Wang, Santos, and
  Xiang}]{shi2021learning}
Peng Shi, Patrick Ng, Zhiguo Wang, Henghui Zhu, Alexander~Hanbo Li, Jun Wang,
  Cicero Nogueira~dos Santos, and Bing Xiang. 2021.
\newblock Learning contextual representations for semantic parsing with
  generation-augmented pre-training.
\newblock \emph{arXiv preprint arXiv:2012.10309}.

\bibitem[{Shiv and Quirk(2019)}]{Shiv2018NovelPE}
Vighnesh~Leonardo Shiv and Chris Quirk. 2019.
\newblock Novel positional encodings to enable tree-structured transformers.
\newblock In \emph{Advances in Neural Information Processing Systems
  (NeurIPS)}.

\bibitem[{Vaswani et~al.(2017)Vaswani, Shazeer, Parmar, Uszkoreit, Jones,
  Gomez, Kaiser, and Polosukhin}]{vaswani2017}
Ashish Vaswani, Noam Shazeer, Niki Parmar, Jakob Uszkoreit, Llion Jones,
  Aidan~N Gomez, \L~ukasz Kaiser, and Illia Polosukhin. 2017.
\newblock \href
  {http://papers.nips.cc/paper/7181-attention-is-all-you-need.pdf} {Attention
  is all you need}.
\newblock In I.~Guyon, U.~V. Luxburg, S.~Bengio, H.~Wallach, R.~Fergus,
  S.~Vishwanathan, and R.~Garnett, editors, \emph{Advances in Neural
  Information Processing Systems (NeurIPS)}.

\bibitem[{Wang et~al.(2020{\natexlab{a}})Wang, Shin, Liu, Polozov, and
  Richardson}]{rat-sql}
Bailin Wang, Richard Shin, Xiaodong Liu, Oleksandr Polozov, and Matthew
  Richardson. 2020{\natexlab{a}}.
\newblock {RAT-SQL}: Relation-aware schema encoding and linking for
  text-to-{SQL} parsers.
\newblock In \emph{Proceedings of the 58th Annual Meeting of the Association
  for Computational Linguistics (ACL)}.

\bibitem[{Wang et~al.(2020{\natexlab{b}})Wang, Shin, Liu, Polozov, and
  Richardson}]{wang2020}
Bailin Wang, Richard Shin, Xiaodong Liu, Oleksandr Polozov, and Matthew
  Richardson. 2020{\natexlab{b}}.
\newblock {RAT-SQL}: Relation-aware schema encoding and linking for
  text-to-{SQL} parsers.
\newblock In \emph{Proceedings of the 58th Annual Meeting of the Association
  for Computational Linguistics (ACL)}.

\bibitem[{Williams and Zipser(1989)}]{williams1989}
Ronald~J. Williams and David Zipser. 1989.
\newblock \href {https://doi.org/10.1162/neco.1989.1.2.270} {A learning
  algorithm for continually running fully recurrent neural networks}.
\newblock \emph{Neural Computation}, 1(2):270--280.

\bibitem[{Yin and Neubig(2017)}]{yin2017syntactic}
Pengcheng Yin and Graham Neubig. 2017.
\newblock \href {https://doi.org/10.18653/v1/P17-1041} {A syntactic neural
  model for general-purpose code generation}.
\newblock In \emph{Proceedings of the 55th Annual Meeting of the Association
  for Computational Linguistics (ACL)}, pages 440--450, Vancouver, Canada.
  Association for Computational Linguistics.

\bibitem[{Yin and Neubig(2019)}]{yin2019}
Pengcheng Yin and Graham Neubig. 2019.
\newblock \href {https://doi.org/10.18653/v1/P19-1447} {Reranking for neural
  semantic parsing}.
\newblock In \emph{Proceedings of the 57th Annual Meeting of the Association
  for Computational Linguistics (ACL)}.

\bibitem[{Yin et~al.(2020)Yin, Neubig, Yih, and Riedel}]{yin-etal-2020-tabert}
Pengcheng Yin, Graham Neubig, Wen-tau Yih, and Sebastian Riedel. 2020.
\newblock \href {https://doi.org/10.18653/v1/2020.acl-main.745} {{T}a{BERT}:
  Pretraining for joint understanding of textual and tabular data}.
\newblock In \emph{Proceedings of the 58th Annual Meeting of the Association
  for Computational Linguistics}, pages 8413--8426, Online. Association for
  Computational Linguistics.

\bibitem[{Yu et~al.(2020)Yu, Wu, Lin, Wang, Tan, Yang, Radev, Socher, and
  Xiong}]{yu2020grappa}
Tao Yu, Chien-Sheng Wu, Xi~Victoria Lin, Bailin Wang, Yi~Chern Tan, Xinyi Yang,
  Dragomir Radev, Richard Socher, and Caiming Xiong. 2020.
\newblock Grappa: Grammar-augmented pre-training for table semantic parsing.
\newblock \emph{arXiv preprint arXiv:2009.13845}.

\bibitem[{Yu et~al.(2018)Yu, Zhang, Yang, Yasunaga, Wang, Li, Ma, Li, Yao,
  Roman, Zhang, and Radev}]{yu2019}
Tao Yu, Rui Zhang, Kai Yang, Michihiro Yasunaga, Dongxu Wang, Zifan Li, James
  Ma, Irene Li, Qingning Yao, Shanelle Roman, Zilin Zhang, and Dragomir Radev.
  2018.
\newblock \href {https://doi.org/10.18653/v1/D18-1425} {{S}pider: A large-scale
  human-labeled dataset for complex and cross-domain semantic parsing and
  text-to-{SQL} task}.
\newblock In \emph{Proceedings of the 2018 Conference on Empirical Methods in
  Natural Language Processing (EMNLP)}, pages 3911--3921, Brussels, Belgium.
  Association for Computational Linguistics.

\bibitem[{Yu et~al.(2019)Yu, Zhang, Yasunaga, Tan, Lin, Li, Heyang~Er, Pang,
  Chen, Ji, Dixit, Proctor, Shim, Jonathan~Kraft, Xiong, Socher, and
  Radev}]{Yu2019sparc}
Tao Yu, Rui Zhang, Michihiro Yasunaga, Yi~Chern Tan, Xi~Victoria Lin, Suyi Li,
  Irene~Li Heyang~Er, Bo~Pang, Tao Chen, Emily Ji, Shreya Dixit, David Proctor,
  Sungrok Shim, Vincent~Zhang Jonathan~Kraft, Caiming Xiong, Richard Socher,
  and Dragomir Radev. 2019.
\newblock Sparc: Cross-domain semantic parsing in context.
\newblock In \emph{Proceedings of the 57th Annual Meeting of the Association
  for Computational Linguistics (ACL)}, Florence, Italy. Association for
  Computational Linguistics.

\bibitem[{Zelle and Mooney(1996)}]{zelle1996}
John~M. Zelle and Raymond~J. Mooney. 1996.
\newblock Learning to parse database queries using inductive logic programming.
\newblock In \emph{Proceedings of the Thirteenth National Conference on
  Artificial Intelligence (AAAI)}.

\bibitem[{Zettlemoyer and Collins(2005)}]{zettlemoyer2005}
Luke~S. Zettlemoyer and Michael Collins. 2005.
\newblock Learning to map sentences to logical form: Structured classification
  with probabilistic categorial grammars.
\newblock In \emph{Proceedings of the Twenty-First Conference on Uncertainty in
  Artificial Intelligence (UAI)}.

\bibitem[{Zhao et~al.(2021)Zhao, Cao, and Zhao}]{zhao2021gp}
Liang Zhao, Hexin Cao, and Yunsong Zhao. 2021.
\newblock \href {http://arxiv.org/abs/2101.09901} {Gp: Context-free grammar
  pre-training for text-to-sql parsers}.

\end{thebibliography}
\newpage

\appendix

\section{Computing supervision through tree hashing}
\label{app:hashing}
In every decoding step  $t$, we wish to compute for every tree $z_\text{new}$ in the frontier $F_{t+1}$ if $z_\text{new} \in \sZ_t^\text{gold}$. This is achieved using tree hashing.
First, during preprocessing, for every height $t$, we compute the gold hashes $h_t^{\text{gold}}$, the hash values of every sub-tree of  $z^\text{gold}$ of height $t$, in a recursive fashion using a Merkle tree hash \citep{DBLP:conf/crypto/Merkle87}. Specifically, we define:  $$\hash(z) = g(\text{label}(z),\hash(z_l),\hash(z_r))$$
Where $g$ is a simple hash function, $z_l,z_r$ are the left and right children of $z$, and $\text{label}(\cdot)$ gives the node type (such as $\sigma$ and $\Pi$).

During training, in each decoding step $t$, since the hash function is defined recursively, we can compute the frontier hashes using the hash values of the current beam. Then, for every frontier hash we can perform a lookup to check if $\hash(z)\in h_t^{\text{gold}}$. Both the hash computation and lookup are done in parallel for all frontier trees using the GPU.
%  are implemented efficiently to find them. On the frontier, we want to know. For every tree  in the frontier we find, since we have the hash values for the current agenda, and the hash function is defined recursively, we can compute the hash values for the entire frontier in parallel
\section{Examples for Relational Algebra Trees}
\label{app:relalg}

We show multiple examples of relation algebra trees along with the corresponding SQL query, for better understanding of the mapping between the two.

\begin{figure*}
\centering
\scalebox{0.8}{

    \begin{tabular}{l}
      \begin{tikzpicture} \Tree [.$\Pi$  [.$\mathcal{G}_{\text{count}}$ *  ]  [.$\sigma$  [.$\land$  [.$=$ flights.destairport  airports.airportcode  ]  [.$=$ airports.city  'Aberdeen'  ] ]  [.$\times$ flights  airports  ] ] ]\end{tikzpicture} \\
      \begin{tikzpicture} \Tree [.$\Pi$  [.$\kappa$  [.$\kappa$  [.$\mathcal{G}_{\text{count}}$ *  ] ] ]  [.$\sigma$  [.$\land$  [.$=$ flights.destairport  airports.airportcode  ]  [.$=$ airports.city  'Aberdeen'  ] ]  [.$\kappa$  [.$\times$ flights  airports  ] ] ] ]\end{tikzpicture} 
      
    \end{tabular}
    }
    \caption{Unbalanced and balanced relational algebra trees for the utterance \emph{``How many flights arriving in Aberdeen city?''}, where the corresponding SQL query is \texttt{ SELECT COUNT( * ) FROM flights JOIN airports ON flights.destairport = airports.airportcode WHERE airports.city = 'Aberdeen'}.
    }
    \label{fig:fig10}
    % \vspace{-2ex}
\end{figure*}
\begin{figure*}
\centering
\scalebox{0.8}{

    \begin{tabular}{l}
      \begin{tikzpicture} \Tree [.$\lambda$ 1   [.$\tau_{asc}$ transcripts.transcript\_date   [.$\Pi$  [.$\sqcup$ transcripts.transcript\_date  transcripts.other\_details  ] transcripts  ] ] ]\end{tikzpicture} \\
      \begin{tikzpicture} \Tree [.$\lambda$  [.$\kappa$  [.$\kappa$  [.$\kappa$ 1  ] ] ]  [.$\tau_{asc}$  [.$\kappa$  [.$\kappa$ transcripts.transcript\_date  ] ]  [.$\Pi$  [.$\sqcup$ transcripts.transcript\_date  transcripts.other\_details  ]  [.$\kappa$ transcripts  ] ] ] ]\end{tikzpicture}
      
    \end{tabular}
    }
    \caption{Unbalanced and balanced relational algebra trees for the utterance \emph{``When is the first transcript released? List the date and details.''}, where the corresponding SQL query is \texttt{ SELECT transcripts.transcript\_date , transcripts.other\_details FROM transcripts ORDER BY transcripts.transcript\_date ASC LIMIT 1}.
    }
    \label{fig:fig11}
    % \vspace{-2ex}
\end{figure*}
\begin{figure*}
\centering
\scalebox{0.8}{

    \begin{tabular}{l}
      \begin{tikzpicture} \Tree [.$\Pi$  [.$\mathcal{G}_{\text{count}}$ *  ]  [.$\sigma$  [.$\land$  [.$\land$  [.$=$ student.stuid  has\_pet.stuid  ]  [.$=$ has\_pet.petid  pets.petid  ] ]  [.$\land$  [.$=$ student.sex  'F'  ]  [.$=$ pets.pettype  'dog'  ] ] ]  [.$\times$  [.$\times$ student  has\_pet  ] pets  ] ] ]\end{tikzpicture} \\
      \begin{tikzpicture} \Tree [.$\Pi$  [.$\kappa$  [.$\kappa$  [.$\kappa$  [.$\mathcal{G}_{\text{count}}$ *  ] ] ] ]  [.$\sigma$  [.$\land$  [.$\land$  [.$=$ student.stuid  has\_pet.stuid  ]  [.$=$ has\_pet.petid  pets.petid  ] ]  [.$\land$  [.$=$ student.sex  'F'  ]  [.$=$ pets.pettype  'dog'  ] ] ]  [.$\kappa$  [.$\times$  [.$\times$ student  has\_pet  ]  [.$\kappa$ pets  ] ] ] ] ]\end{tikzpicture} 
      
    \end{tabular}
    }
    \caption{Unbalanced and balanced relational algebra trees for the utterance \emph{``How many dog pets are raised by female students?''}, where the corresponding SQL query is \texttt{ SELECT COUNT( * ) FROM student JOIN has\_pet ON student.stuid = has\_pet.stuid JOIN pets ON has\_pet.petid = pets.petid WHERE student.sex = 'F' AND pets.pettype = 'dog'}.
    }
    \label{fig:fig12}
    % \vspace{-2ex}
\end{figure*}
\begin{figure*}
\centering
\scalebox{0.8}{
    
    % \begin{center}
    \begin{tabular}{l}
      \begin{tikzpicture} \Tree [.$\Pi$  [.$\mathcal{G}_{\text{count}}$  [.$\delta$ matches.loser\_name  ] ] matches  ]\end{tikzpicture} \\
      \begin{tikzpicture} \Tree [.$\Pi$  [.$\mathcal{G}_{\text{count}}$  [.$\delta$ matches.loser\_name  ] ]  [.$\kappa$  [.$\kappa$ matches  ] ] ]\end{tikzpicture} \\
      
    \end{tabular}
    % \end{center}
    }
    \caption{Unbalanced and balanced relational algebra trees for the utterance \emph{``Find the number of distinct name of losers.''}, where the corresponding SQL query is \texttt{ SELECT COUNT( DISTINCT matches.loser\_name ) FROM matches}.
    }
    \label{fig:fig13}
    % \vspace{-2ex}
\end{figure*}
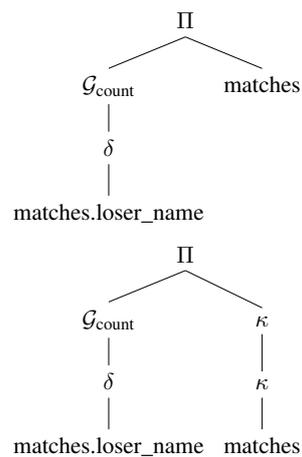

\end{document}